# Efficient Estimation of the Value of Information in Monte Carlo Models

Tom Chávez [1,2] and Max Henrion [1,3]

[1] Rockwell International Science Lab, 444 High St., Palo Alto, CA 94301

[2] Department of Engineering-Economic Systems, Stanford University

[3] The Institute for Decision Systems Research, Inc., 350 Cambridge Ave., #380, Palo Alto, CA 94306

*Abstract:* The expected value of information (EVI) is the most powerful measure of sensitivity to uncertainty in a decision model: it measures the potential of information to improve the decision, and hence measures the expected value of the outcome. Standard methods for computing EVI use discrete variables and are computationally intractable for models that contain more than a few variables. Monte Carlo simulation provides the basis for more tractable evaluation of large predictive models with continuous and discrete variables, but so far computation of EVI in a Monte Carlo setting also has appeared impractical. We introduce an approximate approach based on preposterior analysis for estimating EVI in Monte Carlo models. Our method uses a linear approximation to the value function and multiple linear regression to estimate the linear model from the samples. The approach is efficient and practical for extremely large models. It allows easy estimation of EVI for perfect or partial information on individual variables or on combinations of variables. We illustrate its implementation within Demos (a decision modeling system), and its application to a large model for crisis transportation planning.

## 1.0 EVI: What's so, and What's New

Any model is inevitably a simplification of reality, and most of its input quantities are invariably uncertain. Sensitivity analysis identifies which sources of uncertainty in a model affect its outputs most significantly. In this way, it helps a decision maker focus attention on what assumptions really matter. It also helps a decision modeler to assign priorities to his efforts to improve, refine, or extend his model by identifying those variables for which it will be most valuable to find more complete data, to interview more knowledgeable experts, or to build more elaborate submodels.

The **expected value of information (EVI)** on a variable $x_i$ measures the expected increase in value $y$ if we learn new information about $x_i$ and make a decision with higher-expected value in light of that information. It is the most powerful method of sensitivity analysis because it analyzes a variable's importance in terms of the overall prescription for action, and it expresses that importance in the utility or value units of the problem. Other methods, such as rank-order correlation, express importance in terms of the correlation between an uncertain variable and the output of the decision model. There are many cases where a variable can show high sensitivity in this way, yet still have no effect on the selection of an optimal decision. Deterministic perturbation measures importance in utility or value units, but it ignores nonlinearities and interactions among variables, and also fails to measure a variable's importance in terms of that variable's ability to change the recommended decision.

One calculates **EVPI (Expected Value of Perfect Information)** in discrete models by rolling back the decision tree. The computation itself is straightforward in the sense that, to compute EVI, one simply places at the front of the tree the chance variables to be observed. The EVPI is computed as the difference between the expected value computed for this scenario and the expected value for the regular tree, without observations.

Computing EVI with continuous variables is less intuitive, because we have no tidy way of reversing the uncertainty, unlike the discrete case. Yet continuous models are increasingly the norm for risk and decision analysis, first because discretizing inherently continuous variables introduces unnecessary approximation, and second because Monte Carlo methods and their variants (e.g., Latin hypercube) generate tractable, highly efficient solutions to predictive models that contain thousands of variables. An especially useful feature of the Monte Carlo method is that, for a specified error, the computational complexity increases linearly in the number of uncertain variables [Morgan and Henrion, 1990]. Exact methods require computation time that is exponential in the number of variables.

There is thus a need to develop flexible, efficient methods for computing EVI on continuous variables in a Monte Carlo setting. A **flexible** method has (1) the ability to compute EVI on single variables or on any combination of variables, and (2) the ability to compute both perfect and partial values of information. **Perfect information** removes uncertainty entirely. **Partial information** reduces uncertainty.

We present a general framework for calculating EVI based on preposterior analysis. Using that framework, we develop a technique for computing EVI that depends on a linear approximation to the value function and on multiple linear regression to estimate the constants for the linear function. We also discuss a heuristic method for measuring the value of partial information in terms of what we call the **relative information multiple (RIM)**. We have implemented these methods in detachable computational modules using Demos, a decision modeling system from Lumina, Inc., Palo Alto, CA. We demonstrate their use on a large model to aid in military transportation crisis planning.

## 2.0 Framework

A decision model consists of a set of $n$ state variables $x_1,...,x_n$, which we will denote by **X**. The decision maker has control of a decision variable **D**, which can assume one of $m$ possible values $d_1,...,d_m$. The value or utility function $v(\mathbf{X},d_i)$ expresses the payoff to the decision maker when **X** obtains and decision $d_i$ is chosen.

In a typical decision model, the state variables are uncertain. We express prior knowledge about **X** in the form of a probability distribution, denoted $\{\mathbf{X}|\xi\}$, where $\xi$ denotes a prior state of information. The optimal Bayes' decision maximizing the expected value[1] is given by

$$d^* = \underset{d}{\text{Arg max}} \ \langle v(\mathbf{X},d) | \xi \rangle.$$

The optimal decision given perfect information on state variable $x$, denoted $d^*_x$, is

$$d^*_x = \underset{d}{\text{Arg max}} \ \langle v(\mathbf{X},d) | x,\xi \rangle$$

We define **EVPI** on $x$ as

$$\text{EVPI}\langle x \rangle = \langle v(\mathbf{X},d^*_x) | \xi \rangle - \langle v(\mathbf{X},d^*) | \xi \rangle.$$

In a similar fashion, we define the optimal expected-value decision given the revelation of evidence $e$, $d^*_e$, as

$$d^*_e = \underset{d}{\text{Arg max}} \ \langle v(\mathbf{X},d) | e,\xi \rangle$$

Then the EVI for evidence $e$ is
$$EVI(e) = \langle v(\mathbf{X},d^*_e) | \xi \rangle - \langle v(\mathbf{X},d^*) | \xi \rangle$$

### 2.1 Binary decisions and Function z

Let us consider a simplified decision problem with two decision alternatives: one of them is the optimal Bayes' decision $d^*$; the other we denote $d^+$.

In view of the uncertainty in the state variables, there must exist uncertainty in the outputs as well. Thus, for each decision $d_i$, there exists a unique probability distribution on value $\{v(\mathbf{X},d_i)|\xi\}$ (see Figure 1). For notational convenience we let

$$\bar{v}(d) = \langle v(\mathbf{X}, d) | \xi \rangle.$$

---

[1]. We use Howard's inferential notation (see, for example, Howard,1970). $\{X|S\}$ denotes the probability density of X conditional on S; $\langle X| S \rangle$ denotes the expectation of X conditional on S.

We now define

$$z = v(\mathbf{X}, d^*) - v(\mathbf{X}, d^+).$$

Function $z$ is the pivotal element in our framework for computing EVI because it describes the difference in value between the best and second-best decisions. In Figure 2, we have graphed the probability distribution of $z$. The shaded area represents the total probability of making a bad decision, i.e., doing $d^*$ when $d^+$ would yield higher value. Exploiting information encoded in the shaded, negative portion of the $z$ distribution's curve will provide the necessary clues to compute EVPI and EVI.

**FIGURE 1.** Probability distributions on value for the two decisions $d^*$ and $d^+$.

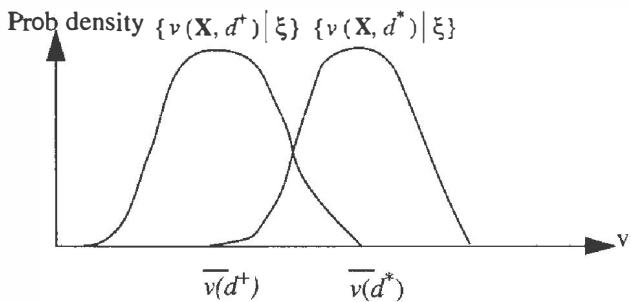

**FIGURE 2.** Function $z$: the difference in value between the best and second-best decisions.

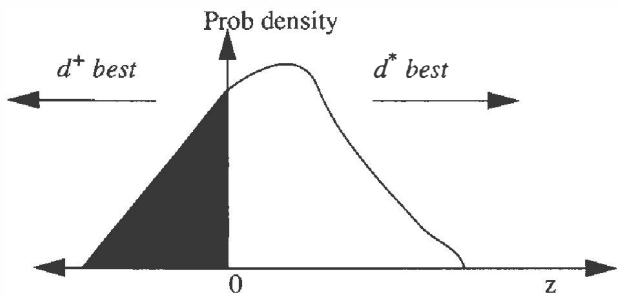

In fact, we can use the intuition behind Figure 2 to write an expression for the **general EVPI**, which is EVPI on all state variables. The absolute value of $z$ in the negative shaded portion is the utility that we could gain by choosing $d^+$ instead of $d^*$; its probability is just its corresponding value on the density curve. Therefore, we have that

$$\text{EVPI} = \int_{-\infty}^{0} |z| \, \{z|\xi\} \, dz. \qquad \text{(EQ 1)}$$

### 2.2 Preposterior Analysis

Preposterior analysis helps us to calculate the effect on $X$ of our seeing evidence $e$, given a prior state of information $\xi$. At the heart of preposterior analysis is the specification of a **preposterior distribution**, which is a a prior probability distribution on a posterior mean. Probability theory provides a principled basis for calculating a preposterior distribution, given a prior and an adequate means of specifying the effects of learning new information.

How do we represent perfect information on a continuous random variable $X$? If $X$ were known with certainty, then its variance would be equal to 0. Thus, we can think of evidence $e$ as an information-gathering activity that somehow reduces the variance of $X$ to 0. Evidence $e$ that provides partial information reduces the variance on the prior of $X$, without shrinking that prior to 0.

The following lemma, taken from basic probability theory, is known as the **conditional expectation formula**.

Lemma 1:   $\langle X|\xi \rangle = \langle \langle X|e \rangle | \xi \rangle$.

A further useful result is the following lemma, which gives the formula for conditional variance.

Lemma 2: $\mathbf{Var}(\langle X|e,\xi \rangle) = \mathbf{Var}(X|\xi) - \langle \mathbf{Var}(X|e)|\xi \rangle$.

Let $\mu'_z$ and $\sigma'^2_z$ denote the prior mean and variance of $z$. They are computed from our prior uncertainties $\mathbf{X}$ and our value function $v$. If we observe $e$, then we might ask how $e$ influences $z$; in particular, we would like to know how $e$ affects $\mu'_z$. We will denote the posterior mean of $z$ given evidence $e$ by $\mu''_z$. The distribution $\{\mu''_z|\xi\}$ is a prior density on the posterior mean $\mu''_z$; that is, it is a preposterior density. Substituting $\mu''_z$ for the inner $\langle X|e \rangle$ on the right-hand side of the equation in Lemma 1 reveals that

$$\langle \mu''_z \rangle = \mu'_z. \qquad \text{(EQ 2)}$$

Eq. 2 shows that the mean of the preposterior distribution $\{\mu''_z|\xi\}$ is the same as the prior mean $\mu'_z$.

If $\sigma''^2_z$ denotes the posterior variance of $z$ after $e$, then application of Lemma 2 shows that

$$\mathbf{Var}\{\mu''_z|\xi\} = \sigma'^2_z - \sigma''^2_z. \quad \text{(EQ 3)}$$

If the prior and posterior on $z$ are normal, then, as proved in [Raiffa and Schlaifer, 1961], the preposterior on $z$ is normal also. That is, the normal distribution is conjugate to the normal sampling process. We thus require $z$ to be distributed normally.

In Figure 3, we show a prior on $z$, a possible posterior on $z$ given evidence $e$, and a preposterior density $\{\mu''_z|\xi\}$. Note that the preposterior has the same mean as the prior, and that its variance is the difference between the prior variance and the posterior variance.

**FIGURE 3.** Prior, posterior, and preposterior densities.

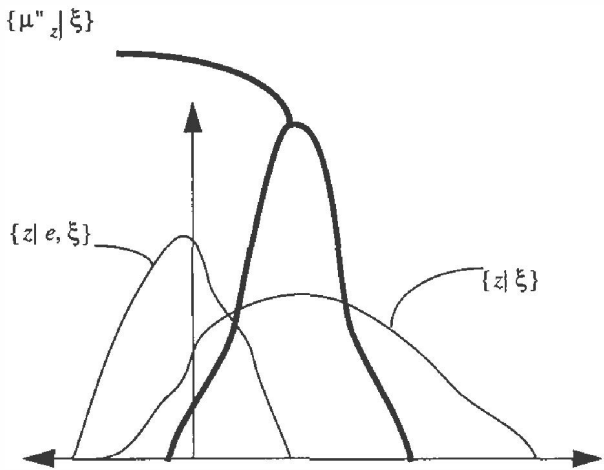

The preposterior density encodes a state of knowledge about $z$ in light of what evidence $e$ might reveal. Its interpretation is the same as in Figure 2. Because it is a probability density on value, we can integrate over its negative area to calculate the EVI of evidence $e$. Thus, we have

$$\mathbf{EVI}(e) = \int_{-\infty}^{0} |\mu''_z| \{\mu''_z|\xi\} d\mu''_z. \quad \text{(EQ 4)}$$

The type of integral given in Eq. 4 is known as a **linear loss integral**. In general, such an integral is impossible to evaluate analytically, so we must rely on statistical tables or numerical approximatation methods to evaluate it.

## 3.0 Complexity and Non-Additivity of EVI

Inference in probabilistic models with discrete variables is exponential in the number of variables, so we would expect the exact calculation of EVI to be exponential in the number of variables also. For simplicity, we assume a single decision variable with $m$ alternatives. Let $k_i$ be the number of states for the $i$th state variable $x_i$. To evaluate a decision tree with $n$ state variables, we require a number of value computations at the leaves equal to

$$m \prod_{i=1}^{n} k_i.$$

Computing EVI on some subset of variables requires at least the same number of value computations at the leaves, and thus we see that exact calculation of EVI is exponential in $n$. Also note that EVI calculations in discrete models are possible for perfect information only.

If $c_i$ represents the value of information on state variable $x_i$, and $C$ represents the value of information on all the state variables simultaneously, then

$$C \neq \sum c_i.$$

The above relation makes it difficult to devise separable, or incremental, procedures for computing EVI, because EVI will often demonstrate nonlinearities for varying combinations of variables and for varying cases of perfect and partial information.

## 4.0 Approximation of EVI

We are ready to apply the preceding analysis to develop an efficient algorithm for estimating EVI. We introduce a linear approximation to the value function, which in turn allows us to derive an expression for $z$, the net difference in value between two decision alternatives. Preposterior analysis on $z$ provides a flexible mechanism for estimating EVI.

## 4.1 The Linear Value Model

We require a key approximating assumption: **The value function $v(X, d_i)$ can be approximated by a first-order (linear) equation for each decision $d_i$**, that is, we can write $v_j$ as a linear function of the $x_i$,

$$v(X, d_j) \approx \sum_j \beta_{ij} x_i + \alpha_j \qquad i = 1...n, j = 1, 2.$$

We assume, for now, that the $x_i$ are independent. (The assumption is not necessary; we use it to simplify our presentation.) We denote the prior mean of $x_i$ by $\mu'_i$ and denote its prior variance by $\sigma'^2_i$. Our approximating assumption allows us to perform simple but useful probabilistic analysis. First, by linearity of expectation, we can write the mean $\bar{v}(d_i)$ as

$$\bar{v}(d_i) = \sum_i \beta_{ij} \mu'_i + \alpha_j. \qquad \text{(EQ 5)}$$

Second, the variance of $\{v(X, d_i)\}$ can be written as

$$Var\{v(X, d_i) | \xi\} = \sum_i \beta_{ij}^2 \sigma'^2_i. \qquad \text{(EQ 6)}$$

By our approximating assumption, we can write a linear approximation for $v(X, d^*)$,

$$v(X, d^*) = \sum_i \beta_i^* x_i + \alpha^*; \qquad \text{(EQ 7)}$$

and one for $v(X, d^+)$,

$$v(X, d^+) = \sum_i \beta_i^+ x_i + \alpha^+. \qquad \text{(EQ 8)}$$

Combining Eqs. 5-8 with the definition of $z$, we can write expressions for the prior mean and variance of $z$:

$$\mu'_z = \overset{\bar{v}(d^*) - \bar{v}(d^+)}{\sum_i (\beta_i^* - \beta_i^+) \mu'_i + (\alpha^* - \alpha^+)}. \qquad \text{(EQ 9)}$$

$$\sigma'^2_z = \sum_i (\beta_i^* - \beta_i^+)^2 \sigma'^2_i. \qquad \text{(EQ 10)}$$

Suppose that $e$ expresses perfect information on $x_k$ and no information about the other $x_i$. Let $\sigma''^2_k$ denote the posterior variance on $x_k$. Since $e$ is perfect information on $x_k$, $\sigma''^2_k = 0$. Eq. 10 gives an approximation to the prior variance on $z$, $\sigma'^2_z$. Given $e$, we know that the $k$th term in the expression in Eq. 10 must be equal to zero. We can thus write the posterior variance for $z$ given $e$, $\sigma''^2_z$:

$$\sigma''^2_z = \sum_{i \neq k} (\beta_i^* - \beta_i^+)^2 \sigma'^2_i. \qquad \text{(EQ 11)}$$

In view of Eq. 3, the preposterior variance on $z$ is

$$Var[\mu''_z | \xi] = (\beta_k^* - \beta_k^+)^2 \sigma'^2_k. \qquad \text{(EQ 12)}$$

## 4.2 Monte Carlo methods: Estimation of the Coefficients

In Monte Carlo simulation, we generate a sample of $n$ scenarios by sampling from the prior distribubtions $\{X|\xi\}$. A scenario $X_s$ is an $n$-tuple of state — variable assignments to $X$. $v(X_s, d_i)$ is equal to the value or utility generated by the $s$th scenario for the $i$th decision alternative.

We can estimate the expected value of each decision $d_i$ as the average of the values $v(X_s, d_i)$ over the scenario index $s$. The optimal Bayes' decision is the maximum of those averages. (Naturally, higher sample sizes give answers of greater precision.) We represent this process for our binary decision problem inTable 1:

**TABLE 1.** Determining the optimal decision in Monte Carlo decision analysis with sample size = 100 and two decision alternatives

| $s$th Scenario | Value with $d_1$ | Value with $d_2$ |
|---|---|---|
| $X_1$ | $v(X_1, d_1)$ | $v(X_1, d_2)$ |
| ⋮ | ⋮ | ⋮ |
| $X_{100}$ | $v(X_{100}, d_1)$ | $v(X_{100}, d_2)$ |
| Average | $\sum_{s=1}^{100} \dfrac{v(X_s, d_1)}{100}$ | $\sum_{s=2}^{100} \dfrac{v(X_s, d_2)}{100}$ |

$$d^* = \arg\max_{j=1,2} \left( \sum_{s=1}^{100} \frac{v(X_s, d_j)}{100} \right)$$

The only outstanding task is to estimate the constants for the linear—approximation model. To this end, we apply multiple linear regression analysis to estimate the constants in Eqs. 7 and 8. Let $i$ be an index into set of $m$ decision alternatives, and let $j$ be an index into the $n$ state variables. From [Shavelson, 1988], we can use multiple linear regression to write constants for $v_i$ — value for the $i$th decision alternative in terms of the $n$ state variables—as follows:

$$\beta_{ij} = \frac{R_{ij} - \sum_{k \neq j} R_{ik} r_{ji}}{1 - \sum_{k \neq j} r^2_{ji}} \frac{\sigma_i}{\sigma_j}, \quad \text{(EQ 13)}$$

where $R_{ij} =$ correlation$(v_i, x_j)$, $R_{ij} =$ correlation$(x_i, x_j)$, $S_i =$ standard deviation$(v_i)$, and $\sigma_i =$ standard deviation$(x_i)$. We estimate these quantities directly from our Monte Carlo samples.

Recall from Section 2.1 that the $v_i$ generate probability distributions in a Monte Carlo model. Thus, it makes sense to think of them as random variables with corresponding sample correlations and standard deviations. The $\alpha_i$ are estimated as follows:

$$\alpha_j = \langle v_j \rangle - \sum_{i=1}^{n} \beta_{ij} \mu'_i.$$

### 4.3  Relative Information Multiple

Suppose now that $e$ expresses partial, rather than perfect, information on $x_k$. It is not immediately obvious how to specify partial information on an uncertain variable. We suggest the following method, based on our concept of a **RIM**. A RIM of evidence $e$ on variable $x_k$ is defined to be the ratio between the prior variance $\sigma'^2_k$ and the posterior variance $\sigma''^2_k$ on $x_k$ after $e$ has been seen. In intuitive terms, the RIM measures how much we *could* know relative to what we know *now*. It is a multiple on missing but knowable information. For example, if an information source could tell me roughly twice as much as I know now, then the equivalent RIM is 2.

A variable $x_k$'s contribution to the prior variance $\sigma'^2_z$ is given in Eq. 10 as $(\beta^*_k - \beta^+_k)^2 \sigma'^2_k$. For a RIM $= r$ of evidence $e$ on variable $x_k$, the posterior variance $\sigma''^2_z$ is given by

$$\sigma''^2_z = \frac{1}{r} (\beta^*_k - \beta^+_k)^2 \sigma'^2_k. \quad \text{(EQ 14)}$$

The preposterior variance is estimated as

$$\mathbf{Var}[\mu''_z | \xi] = \frac{r-1}{r} (\beta^* - \beta^+)^2 \sigma'^2_k. \quad \text{(EQ 15)}$$

The preposterior mean for partial information stays the same, as in Eq. 9.

### 4.4  Z Is Normal

We will assume that the $x_i$ are normally distributed. In light of the following proposition from probability theory, our linear—approximating assumption requires $z$ to be normally distributed also.

**Proposition**: Let $X_i$ be a collection of $n$ normal random variables with means given by $\mu_i$ and variances $\sigma^2_i$. Define the random variable $Y$ as

$$Y = \alpha + \sum_i \beta_i x_i.$$

Then Y is normally distributed also, with mean given by

$$\alpha + \sum_i \beta_i \mu_i,$$

and variance

$$\sum_i \beta^2_i \sigma^2_i.$$

□

Observe that our approximating assumption allows us to write the mean and variance of $z$ using standard probability formulae. There is nothing about our framework, however, that forces the actual distribution of $z$ to belong to the same family as do $z$'s component distributions. For exam-ple, if the $x_i$ are Poisson, normal, and exponential, then $z$ is a hard-to-assess, mongrel distribution. Assuming that the $x_i$ are normal forces $z$ to be normal also. If the $x_i$ are non-normal, then we must make an extra approximating assumption that $z$ is normal also, although we must emphasize that this assumption would not be analytically true.

A limiting aspect of the technique presented here is that it measures EVI relative to only two decisions. In [Chavez, 1994], we show how to extend it to accommodate multiple ($\geq 3$) decision alternatives.

### 4.5 Algorithm for EVI

We now summarize, in algorithmic form, our general technique for estimating EVI in a Monte Carlo decision model:

1. Select the two decision alternatives generating the highest and second-highest expected value, $d^*$ and $d^+$.

2. Define variable $z$ as the difference between $v(\mathbf{X},d^*)$ and $v(\mathbf{X},d^+)$.

3. Calculate regression constants $\beta_i^*$, $\beta_i^+$, $\alpha^*$, and $\alpha^+$.

4. Using Eqs. 2 and 9, calculate the mean $\langle \mu''_z | \xi \rangle$ of the preposterior distribution of $z$:

$$\langle \mu''_z | \xi \rangle = \mu'_z = \sum_j (\beta_j^* - \beta_j^+) \mu'_j + (\alpha^* - \alpha^+).$$

5. For perfect information on $X_i$, define

$$\mathbf{Var}\{\mu''_z | \xi\} = (\beta^* - \beta^+)^2 \sigma'^2_i.$$

6. For partial information on $X_i$ with RIM=$k$, define

$$\mathbf{Var}\{\mu''_z | \xi\} = \frac{k-1}{k} (\beta^* - \beta^+)^2 \sigma'^2_i.$$

7. For perfect information on variables with indices in $S$, define

$$\mathbf{Var}\{\mu''_z | \xi\} = \sum_{j \in S} (\beta^* - \beta^+)^2 \sigma'^2_j.$$

8. For partial information on variables with indices $i$ and a corresponding ordered set of RIM's $k_i$, define

$$\mathbf{Var}\{\mu''_z | \xi\} = \sum_i \frac{k_i - 1}{k_i} (\beta^* - \beta^+)^2 \sigma'^2_i.$$

9. For perfect information on variables with indices in $S$ mixed with partial information on variables with indices $i$ and a corresponding ordered set of RIM's $k_i$,

$$\mathbf{Var}\{\mu''_z | \xi\} = \sum_i \frac{k_i - 1}{k_i} (\beta^* - \beta^+)^2 \sigma'^2_i + \sum_{j \in S} (\beta^* - \beta^+)^2 \sigma'^2$$

10. Define the preposterior density on $z$, $\{\mu''_z | \xi\}$, as

$$\text{Normal}(\langle \mu''_z | \xi \rangle = \mu'_z, \mathbf{Var}\{\mu''_z | \xi\})$$

11. Express EVI as

$$\mathbf{EVI} = \int_{-\infty}^{0} |\mu''_z| \ \{\mu''_z | \xi\} d\mu''_z$$

12. Perform the integration in (11) numerically.

## 5.0 Application

We now describe an application of our method to a large decision model developed at Rockwell's Palo Alto Science Laboratory to support **Course of Action (COA)** analysis for **Noncombatant Evacuation Operations (NEO)**. Implemented in Demos, NEO-COA allows a user to instantiate a generic NEO plan with specific parameter values for locations, forces, and destinations of troops and civilians. The model provides insights into the relative strengths of alternative plans by scoring them using different evaluation metrics, such as time to complete the operation. Because many of the elements of a real-world military planning scenario are not known with certainty, several of the model's inputs are specified as continuous probability distributions.

In the current version of NEO-COA, there are three decision variables, or factors over which a military planner exercises control:

- *Security forces*: Security forces vary in their starting locations, dates of availability, and capabilities in providing security.
- *Safe havens*: The places where civilians gather to take shelter, safe havens differ in terms of distances from the assembly areas and port capacities.
- *Transportation assets*: A configuration of transportation assets is a sequencing of transportation capability over a fixed period of time. "Three C-141's available on day 2 and 5 C-141's on days 3 through 10" is an example of a particular transportation configuration.

Because each of these decision variables currently possess three alternatives, there are a total of 27 available courses of action. In addition, the NEO-COA model possesses over 100 different input variables; of those, currently nine are specified as probabilistic quantities.

Once the decision variables and inputs have been specified, the model performs a dynamic simulation of the flow of U.S. citizens (the non-combatants) from their starting locations within a country to a set of selected assembly areas, and then on to the safe havens. It also includes risk factors associated with both U.S. citizens and U.S. military personnel as functions of time. For example, risk to U.S. citizens at the assembly areas can rise and fall over the course of an entire operation in response to uncertain events, such as the arrival of security forces. The functional representations of the risk factors are then used to compute expected casualties — civilian and military — for varying alternatives.

A top level view of the NEO-COA model as implemented in Demos is shown in Figure 4. There are three uncertainties for the NEO-COA model: **Initial USCITS**, probability distributions on the number of U.S. citizens in each of the three regions of the country (capital, north, and south) at the start of a crisis planning operation; **Country Regions Attrition Risk**, which is the risk posed to noncombatants over the course of an operation; and **Transfer Rate**, which is the speed at which civilians move from their starting locations to the assembly areas. Thus a total of nine continuous probability distributions must be assigned; typically, these are subjective assessments provided by military planners using the model.

In Figure 5, we show the results of applying the EVI approximation technique to the NEO model for perfect information. We see, for example, that the uncertainty about the number of American citizens in the capital has EVI equal to about six lives, and the uncertainty about the transfer rates in the capital has perfect information value equal to more than seven lives. In all cases, the value of information is highest for uncertainties relating to the capital region, reflecting that the highest number of citizens are concentrated there. The integral in Eq. 4 is evaluated numerically. Perfect information calculations on nine uncertainties took Demos 1 minute, 53 seconds running on a Macintosh IIfx computer.

## 6.0 Conclusions and Future Directions

We have described a general analytic framework for estimating EVI in a decision model using preposterior analysis. It employs a linear—approximating assumption that allows us to write the value function as a first-order equation in the inputs. We define variable $z$ to be the difference in value for the two decision alternatives. Multiple linear regression on the inputs provides the necessary constants for the linear value equation; we estimate the regression constants from Monte Carlo sample information. Applying preposterior analysis to $z$ allows us to write an approximation to the value of perfect and partial information for any combination of state variables.

There are several areas in which we plan to extend the work presented here. First, we would like to develop a sister technique for approximating EVI on continuous *decision* variables. Second, we would like to examine how well our technique performs relative to an exact, more costly approach. To this end we will apply our method to several large models, run it several times, and compare its results to the corresponding exact answers. Third, we wilk apply statistical proof techniques to analyze formally the algorithm's convergence and error characteristics.

## 7.0 References

Chávez, Tom. (1994). "Recovering Value of Information from Pairwise Peeks," Rockwell Palo Alto Science Lab, Technical Memorandum.

Henrion, Max and Morgan, Granger. (1990). *Uncertainty: A Guide to Dealing with Uncertainty in Quantitative Risk and Policy Analysis*. Cambridge University Press, Cambridge.

Howard, R. A. (1970). "Proximal Decision Analysis," *Management Science*, 17, No. 9:507-541.

Raiffa, Howard and Schlaifer, Robert. (1961). *Applied Statistical Decision Theory*, MIT Press, Boston.

Shavelson, Richard. (1988). *Statistical Reasoning for the Behavioral Sciences*, 2nd ed, Allyn and Bacon, Inc, Boston.

**FIGURE 4.** Top-level view of the NEO-COA model.

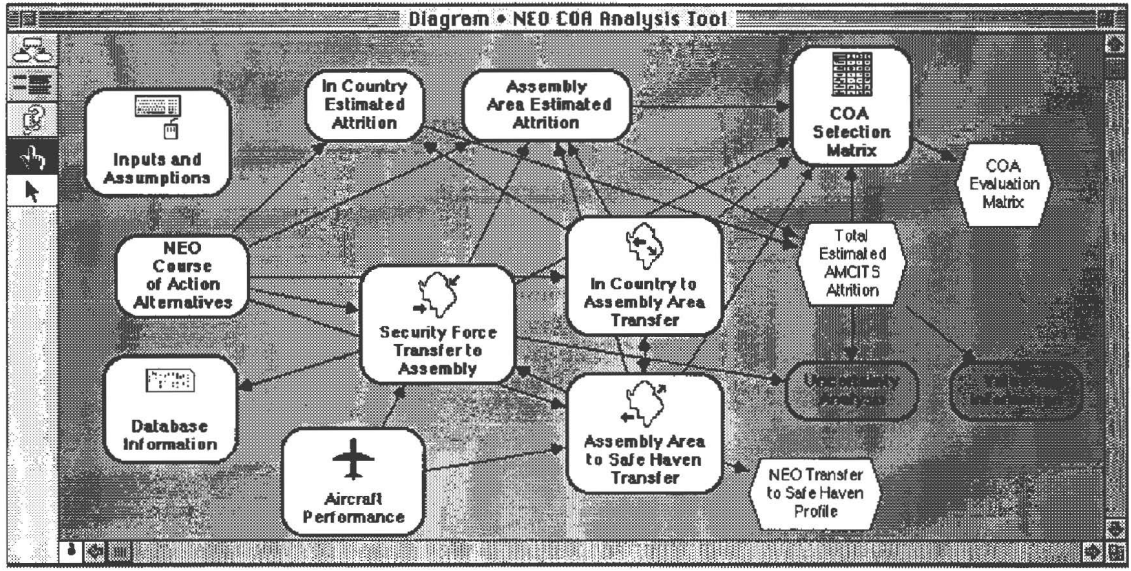

**FIGURE 5.** Approximation of perfect information values.

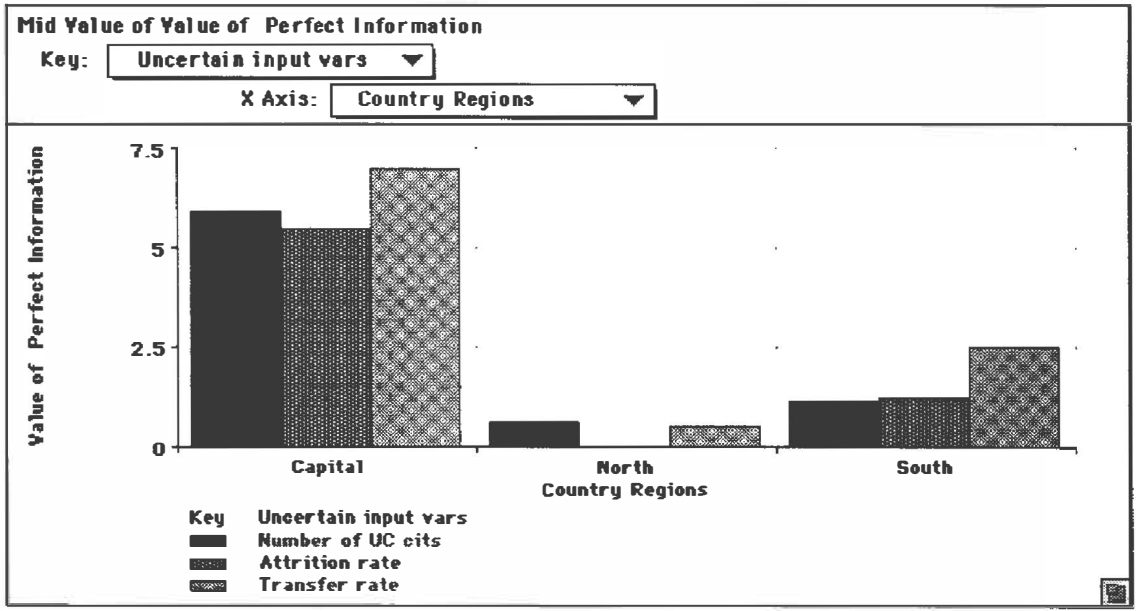